\documentclass[5p,times]{elsarticle}

\usepackage{lineno,hyperref}
\modulolinenumbers[5]

\journal{Journal of \LaTeX\ Templates}

\usepackage{color}
\usepackage{amsfonts}
\usepackage{amsmath}
\usepackage{multirow}

\begin{document}

\begin{frontmatter}

\title{Communication-Efficient Separable Neural Network for Distributed Inference on Edge Devices}



\author{Jun-Liang Lin}
\ead{r06921060@ntu.edu.tw}

\author{Sheng-De Wang\corref{mycorrespondingauthor}}
\cortext[mycorrespondingauthor]{Corresponding author}
\ead{sdwang@ntu.edu.tw}

\address{Department of Electrical Engineering, National Taiwan University, Taipei 10617, Taiwan}




\begin{abstract}
The inference of Neural Networks is usually restricted by the resources (e.g., computing power, memory, bandwidth) on edge devices. In addition to improving the hardware design and deploying efficient models, it is possible to aggregate the computing power of many devices to enable the machine learning models. In this paper, we proposed a novel method of exploiting model parallelism to separate a neural network for distributed inferences. To achieve a better balance between communication latency, computation latency, and performance, we adopt neural architecture search (NAS) to search for the best transmission policy and reduce the amount of communication. The best model we found decreases by 86.6\% of the amount of data transmission compared to the baseline and does not impact performance much. Under proper specifications of devices and configurations of models, our experiments show that the inference of large neural networks on edge clusters can be distributed and accelerated, which provides a new solution for the deployment of intelligent applications in the internet of things (IoT).
\end{abstract}


\end{frontmatter}

\nolinenumbers
\section{Introduction}
In these years, deep learning has been widely used in lots of areas such as computer vision, natural language processing, and audio recognition. To obtain state-of-the-art results, models become much larger and deeper. However, large neural networks are so computationally intensive that the loadings are heavy for general low power edge devices. For example, NoisyStudent \cite{xie2020self} proposed by Xie \textit{et al.} has achieved a state-of-the-art performance of 88.4\% top-1 accuracy on ImageNet. Their model EfficientNet-L2 is extremely large with 480 million parameters, making the model unsuitable for deploying on edge devices. To trade-off between accuracy and latency, many lightweight models such as MobileNet \cite{howard2017mobilenets} and SqueezeNet \cite{iandola2016squeezenet} have been proposed. Many model compression techniques are also widely used \cite{han2015deep}. Nevertheless, the performance of lightweight models is still far from the state-of-the-art models and model compression techniques suffer from dramatic accuracy drop once reaching the limit.

Distributed inference seems to be a solution to take care of both accuracy and latency. With the development of the internet of things (IoT), many edge devices can join a network and compose a cluster. Making good use of the idle devices in the cluster has become an issue and is still under discussion \cite{mao2017survey}. If we can distribute the inference process properly on these devices, we can make the inference of a large model much faster without a performance drop. There are two typical distributed methods usually being considered, data parallelism \cite{krizhevsky2012imagenet} and model parallelism \cite{dean2012large}. However, in many inference scenarios on edge  devices, such as object detection, devices receive streaming data of images and make inferences with a pre-trained model. This kind of data must be input one by one instead of batch by batch, so it lacks the mechanism of data parallelism. In this case, model parallelism seems to be a better solution to deal with this problem. 

\begin{figure}[htbp]
\begin{center}
\includegraphics[width=.9\linewidth]{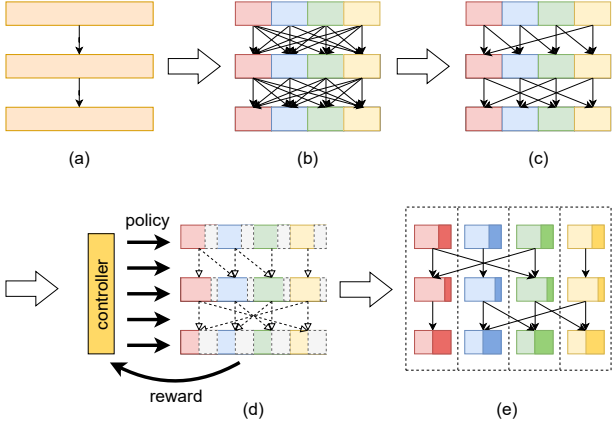}
\end{center}
   \caption{The pipeline of training separable neural network. (a) Original model with a sequence of convolutional layers/blocks. (b) Separating the model straightforwardly cause large transmission overhead. (c) An example of transmission reduction. (d) RL-based neural architecture search to search for best communication policy and sparsification. (e) Fine-tune the best model and deploy for distributed inference. }
\label{fig:paper_pipeline}
\end{figure}

In this paper, we proposed an architecture, Separable Neural Network (SNN), with a new model parallelism method to make distributed inference on edge devices efficiently. With our approach, we can parallelize a class of very deep neural network models, such as ResNet and ResNeXt, to a cluster of edge devices and reduce the costs and overheads of transmission. It not only speeds up the inference but also reduces the memory usage and computation per device. Fig. \ref{fig:paper_pipeline} shows the pipeline of our approach. We separate the original model and introduce an RL-based neural architecture search method to find the best communication policy from an extremely large searching space. Finally, we fine-tune the best model according to decisions made by the policy and deploy it on edge devices. Our method is more suitable for the models with residual connections and group convolutions.

\section{Related Work}
\subsection{Multi-path structures}
Benchmark convolutional neural networks such as AlexNet \cite{krizhevsky2012imagenet}, VGG \cite{russakovsky2015imagenet} are in general designed as a sequence of convolution layers to extract features from low level to high level. GoogLeNet \cite{szegedy2015going} and the series of  Inception networks \cite{szegedy2016rethinking} \cite{DBLP:conf/aaai/SzegedyIVA17} present a concept of parallelism, which extracts features with different size of filters at the same time and merge them together. 

ResNet \cite{he2016deep} introduced residual connections to make features and gradients propagate more unimpededly, and lots of works follow the thought. ResNeXt \cite{xie2017aggregated} is one of the following works. It splits the layers into multi-paths and all the paths can compute in parallel. The multi-path structure can be implemented by group convolution, which divides input channels and output channels into several groups. The number of paths is controlled by cardinality $C$, representing the number of groups. If there are $M$ input channels, $N$ output channels in a group and the cardinality number is $C$, the kernel size is $K$, then the total number of input channels is $C \times M$, and the total number of output channels is $C \times N$. The total number of parameters is $C \times (K \times K \times M \times N)$, which is $C$ times smaller than normal convolution under the same number of input channels and output channels.

\subsection{Model parallelism}


Model parallelism \cite{dean2012large} aims to split the model into several parts and distribute them on different compute nodes. Since each node only computes part of the entire model, it doesn’t need to synchronize the weights and gradients with other nodes. However, this kind of parallelism increases the dependency between the nodes \cite{10.1145/3320060}. That is, one node needs to wait for the results from previous nodes to run the computation. The extra data transmission between layers should also be considered.

\subsection{Distributed inference for neural networks}
Distributed inference is a way to improve inference on edge devices. Teerapittayanon et al. \cite{teerapittayanon2017distributed} proposed a hierarchical model, distributed deep neural network (DDNN), to distribute the components of the model on the cloud, the edge, and the end devices. The end-to-end DDNN can be jointly trained and minimizes transmission costs. DDNN can use the swallow portions in the edge to make inference faster or send the data to the cloud if the local aggregator determines that the information is insufficient to classify accurately. Edgent \cite{li2018edge} is a framework to adaptively partition DNN on edge and end devices according to the bandwidth. After well-trained partitioning, Edgent can make a co-inference on the hybrid resources. It also introduces an early exit mechanism to balance latency and accuracy. DeepThings \cite{zhao2018deepthings} is a framework for adaptively distributed CNN models on edge devices. It can divide convolutional layers into several parts to process and minimize memory usage to reduce communication costs. Furthermore, it can be used in dynamic application scenarios with an adaptive partition method to prevent synchronization overheads.

\subsection{Neural architecture search}
Since rule-based design architectures reach the performance limit, neural architecture search, a method of automatically searching for the best model, becomes more and more popular. However, the searching space of architecture candidates is huge, which makes random searching not practical. Zoph et al. \cite{zoph2016neural} proposed Neural Architecture Search (NAS), which constructs an RNN controller to sample models from searching space and updates the controller based on Reinforcement Learning (RL). This method successfully found a state-of-the-art model but still spent lots of time and computation power. Pham et al. \cite{pham2018efficient} proposed Efficient Neural Architecture Search (ENAS) to improve NAS by sharing models’ weight, omitting the process of training sample models from scratch, using mini-batch data to evaluate performance. Finally, ENAS finds a model with similar performance and achieves 1000 times speedup compared to the original NAS.

\section{Method}
\subsection{Separable Neural Network} \label{seperate}

Started from a model with sequential layers, we denote the feature in layer $l$ as $X^l$. If we separate $X^l$ into $C$ groups, then $X^l$ can be seen as the concatenation of $C$ components.
\begin{equation}
X^l = [X_1^l, X_2^l, ... , X_C^l]
\end{equation}

To get the feature in the next layer, each of the component has to do the following computation:
\begin{equation}
X_i^l = X_i^{l-1} + T_i^{l-1}(X_i^{l-1}) + D_i^l
\end{equation}
where $i$ denotes the $i$-th component, $T$ denotes a transformation such as convolution, and $D$ denotes the feature received from other component. After we get the new feature, a function $F$ with quantization and sparsification further reduces the size of $X$ to make it more efficient for transmission. Another pre-defined function $s$ then decides to send the feature to a set of receivers. We can simply define this process with following equation:
\begin{equation}
D_{s(i, l)}^{l+k} = F(X_i^l)
\end{equation}

\begin{table*}[h!]
\centering
\begin{tabular}{|cc|c|c|}
\hline
stage             & output               & ResNeXt-56 (8$\times$16d)  & Sep-ResNeXt-56 (8$\times$16d) \\ \hline
conv0             & 32x32                & 3x3, 16, stride 1   & 3x3, 16, stride 1            \\ \hline
                  & 32x32                & -                   & concatenate 4 duplications   \\ \hline
conv1             & 32x32                & 
$
\begin{bmatrix}
1\times1,16\\ 
3\times3,128,C=8\\ 
1\times1,64
\end{bmatrix}\times6  $                  &
$\begin{bmatrix}
1\times1,64,C=4\\ 
3\times3,128,C=8\\ 
1\times1,256,C=4
\end{bmatrix}\times6  $                               \\ \hline
conv2             & 16x16                &
$\begin{bmatrix}
1\times1,64\\ 
3\times3,256,C=8\\ 
1\times1,128
\end{bmatrix}\times6  $                  &
$\begin{bmatrix}
1\times1,256,C=4\\ 
3\times3,256,C=8\\ 
1\times1,512,C=4
\end{bmatrix}\times6  $                               \\ \hline
conv3             & 8x8                  & 
$\begin{bmatrix}
1\times1,128\\
3\times3,512,C=8\\ 
1\times1,256
\end{bmatrix}\times6  $                  &
$\begin{bmatrix}
1\times1,512,C=4\\ 
3\times3,512,C=8\\ 
1\times1,1024,C=4
\end{bmatrix}\times6  $                               \\ \hline
                  & 8x8                  & -                   & keep first 1/4 channels      \\ \hline
 & \multirow{2}{*}{1x1} & global average pool & global average pool          \\ \cline{3-4} 
                  &                      & 100-d fc, softmax   & 100-d fc, softmax            \\ \hline
\multicolumn{2}{|c|}{\# of params.}         & 4.39M               & 4.54M                        \\ \hline
\end{tabular}
\caption{Overview of model architectures. (Left) ResNeXt-56. (Right) Sep-ResNeXt-56. We follow the expressions in ResNet and ResNeXt, which $C$ indicates the number of groups in group convolution.}
\label{table:architecture_compare}
\end{table*}

\paragraph{Analysis} \label{analysis}

With a little modification in the training stage, the model becomes separable during inference. The original ResNet \cite{he2016deep} bottleneck block has: 
\begin{equation}
M \cdot N+K \cdot K \cdot N \cdot N+N \cdot M 
\end{equation}
parameters, where M is the number of input channels, N is the output channels and K is the filter size. ResNeXt \cite{xie2017aggregated} further modifies the blocks into multi-path structures, and the number of parameters become:
\begin{equation}
\begin{split}
& C \cdot (M \cdot d+K \cdot K \cdot d \cdot d+d \cdot M), 
\end{split}
\label{eq:2}
\end{equation}
where the number of paths $C$ and the number of channels in middle layer $d$ can be adjusted to make amount of parameters as close as possible to the original ResNet for comparison. We can further rewrite Equation \ref{eq:2} as follows:
\begin{equation}
\begin{split}
& M \cdot (C \cdot d)+C \cdot (K \cdot K \cdot d \cdot d)+(C \cdot d) \cdot M, 
\end{split}
\label{eq:3}
\end{equation}
It is clear to see that the first layer of the block has $M$ input channels, $C\cdot d$ output channels, and the middle layer is a group convolutional layer with $C\cdot d$ input channels, $C\cdot d$ output channels and $C$ convolutional groups.

To construct Separable ResNeXt, we first decide $G$, the maximum number of compute nodes used to deploy the distributed inference system on, then we divide the $C$ paths into $G$ groups. Equation \ref{eq:2} becomes 
\begin{equation}
\begin{split}
& G \cdot (C/G) \cdot (M \cdot d+K \cdot K \cdot d \cdot d+d \cdot M)
\end{split}
\label{eq:4}
\end{equation}
Again, we rewrite Equation \ref{eq:4} as follows:
\begin{equation}
\begin{split}
& G \cdot (M \cdot (C/G) \cdot d)+ C \cdot (K \cdot K \cdot d \cdot d) + \\
& G \cdot  ((C/G) \cdot d \cdot M)
\end{split}
\end{equation}
Comparing with Equation \ref{eq:3}, the first layer becomes group convolutional layer, so the number of convolutional group is $G$ and the number of input channels becomes $G \cdot M$. The middle layer is exactly the same. As a result, the total number of parameters is theoretically the same as ResNeXt. However, since every group has to maintain its own batch normalization and downsampling operations, the size of separable models may increase by about $1\%\sim6\%$.

Table \ref{table:architecture_compare} compares ResNeXt-56 and Sep-ResNeXt-56, and their corresponding amount of parameters, where Sep-ResNeXt-56 denotes separable ResNeXt-56. The most different is that in Sep-ResNeXt we concatenate $G$ replications of the output feature maps of the first layer when training to simulate the synchronization of data on $G$ devices in the first transmission step when deployment. Additionally, to avoid large transmission overhead when aggregating all the results on all devices, we tend to make the device use local feature maps to run classification. Therefore, we only keep the first $1/G$ feature maps in the last convolutional layer, and these feature maps become the input to the final classifier.

The setting of $G$ is related to the number of compute nodes and gives some flexibility of deployment. For example, if we set $G$ as 4, we expect the model to be used in a cluster with four devices. Additionally, the model can also be deployed on less than four compute nodes. We can put four parts together or put two parts on two compute nodes, respectively. These scenarios of deployment perform the same on accuracy.

\begin{figure*}[h!]
\begin{center}
\includegraphics[width=.8\linewidth]{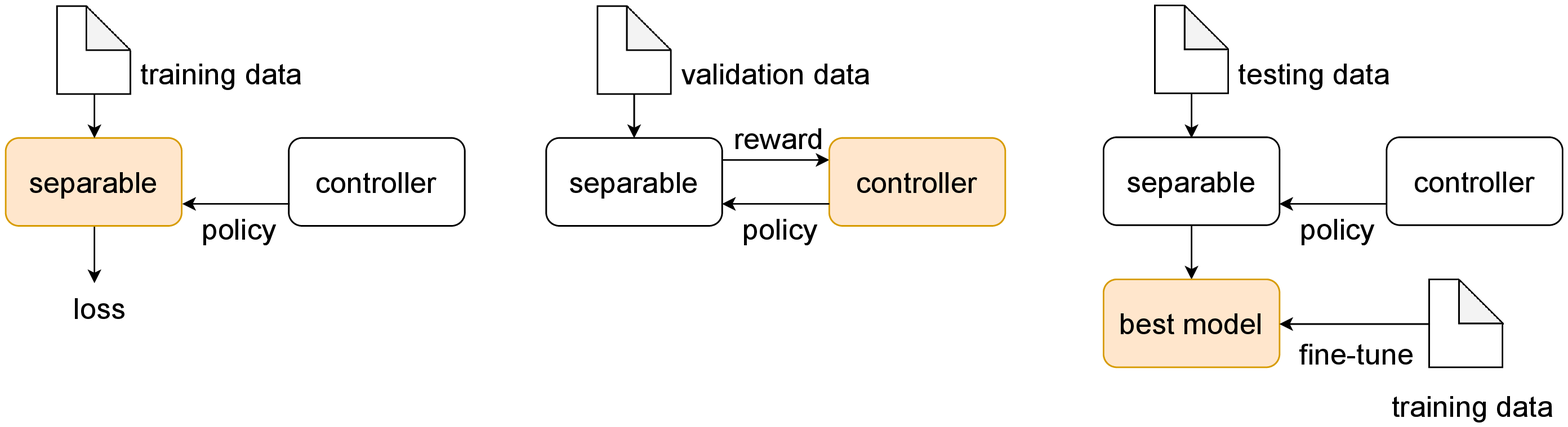}
\end{center}
   \caption{The overview of searching for the best model with NAS. (Left) Train the separable network. (Middle) Train the controller network. (Right) Sample the best model and fine-tune. }
\label{fig:enas_flow}
\end{figure*}

\subsection{Neural Architecture Search for the best communication policy} \label{enas_based}

After we separate the model, parts of the model are detached from each other. If all the parts transmit their information to each other, there would be a large amount of data transmission and cause large overheads. We make use of neural architecture search (NAS) to deal with this problem. With the method of neural architecture search, we can just keep the transmissions that are critical for the model's performance. 

The search method contains two components: a separable network and a controller network. We use a computational graph to represent a neural network model -- the nodes are the convolutional layers of the model, and the edges denote the data flow. In this perspective, the separable network contains the nodes of the graph and the controller network control the edges. Since the edges determine the transmissions, every decision sampled from the controller in a step determines a transmission scheme. The controller is only used in the training stage. After getting the best model from the controller, we no longer use it in the inference stage. 

Since the models are encoded as sequences and recurrent neural network (RNN) is suitable for dealing with the sequence data, we construct the controller network using RNN. In the first step, the RNN controller uses default input and default hidden state to generate an output, representing every possible communication's probability. We sample a decision in this step according to the probability. The decision and the current hidden state then become the input and hidden state to the next step. We repeat the process until all the decisions have been made. Furthermore, we expect that every compute node only deals with one sending process and one receiving process, so the destination of data to be sent should not be duplicated. Based on this premise, the number of choices of each step can be specified as $G!$, where $G$ is the number of compute nodes. 

The training process can be divided into three stages: (1) training the separable network, (2) training the controller network, (3) sampling the best model from the controller, and fine-tuning. Fig \ref{fig:enas_flow} shows the overview of these three stages. We use different sets of data to train different networks in different stages. In the following, we will describe the details in order.

\begin{enumerate}[(1)]
\item{Training the separable network} \\
To train the separable network, we repeat the following steps: first, we fix the parameter $ \theta_{\pi} $ of controller network $\pi$, then sample decisions from the policy output by the controller to determine a model $m$ with the parameters of separable network $\theta_{s}$. The expected loss $\mathbb{E}_{m\sim\pi}[\mathcal{L}(m;\theta_{s})]$ can be estimated by the Monte Carlo method. According to the expected loss, gradients can be computed as
\begin{equation}
\bigtriangledown_{\theta_{s}}\mathbb{E}_{m\sim\pi}[\mathcal{L}(m;\theta_{s})]\approx\frac{1}{M}\sum_{i=1}^{M}\bigtriangledown_{\theta_{s}}\mathcal{L}(m;\theta_{s})
\end{equation}
and finally, the separable network can be updated using stochastic gradient descent algorithm. 

\item{Training the controller network}\\
As for controller network $\pi$, we fix the parameters of the separable network $\theta_{s}$, then sample decisions to determine a model $m$. With the sampled model, we make inference on validation data and take the accuracy as reward $\mathcal{R}$. In this stage we try to maximize the expected reward $\mathbb{E}_{m\sim\pi}[\mathcal{R}]$ by using reinforcement learning. We can compute the gradient with the policy gradient method and update parameters $ \theta_{\pi} $.

\item{Sampling the best model from the controller network}\\
After the training on separable network and controller network, we simply sample some models with the method in the training stage to measure the performance on both two networks and find the best models. We keep the best weights and decisions of the separable network model every epoch and fine-tune with a small learning rate to get the best result.

\end{enumerate}

\subsection{Transmission overhead reduction} \label{reduction}
\begin{figure*}[htbp]
\begin{center}
\includegraphics[width=.8\linewidth]{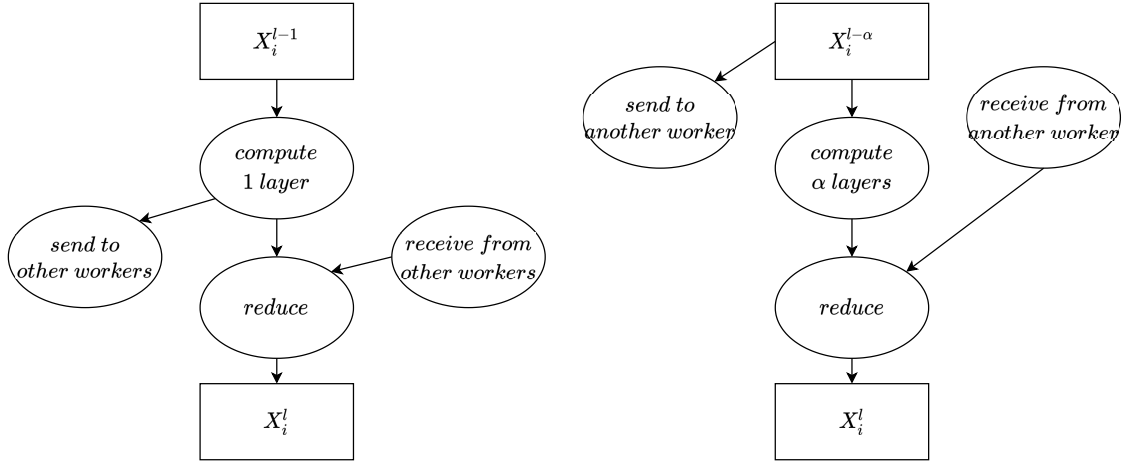}
\end{center}
   \caption{The differences of computation graph before and after introducing the staleness factor $\alpha$. (Left) In fully synchronous scenario, every worker has to send its output to other workers after computing each layer. The computation and transmission can not be done in parallel. (Right) With staleness factor $\alpha > 1$, we can proceed the computation and the transmission at the same time.}
\label{fig:computation_graph}
\end{figure*}

To make the process perform computation and transmission at the same time, we use a staleness factor denoted by $\alpha$ to control the tolerant delay between sending and receiving. The staleness factor $\alpha$ can be defined as the ratio the of number of block computing to block transmission. In default, the staleness factor $\alpha$ is assumed to 1, which means one block computing and one transmission. If we let the staleness factor $\alpha > 1$,  during transmission, compute nodes do not wait for the data arrived but keep computing. After the transmission done, compute nodes finally aggregate the current result with the data they received. So in the optimal scenario, the transmission time can fully overlap with the computation time. If we set $\alpha$ to 2, it means that on average there would be one transmission every two blocks being computed, and it would save half of the transmission. Fig. \ref{fig:computation_graph} illustrates the concept of the staleness factor, where With staleness factor $\alpha > 1$, we can do more block computations and save the transmission at the same time.

Moreover, we try to add a sparsification decision to control how much data need to be sent at every transmission. We set $K$ levels between $P_{min}\%\sim$100\% to control the percentage of feature maps to be sent. So the amount of data to be sent can be decided by following equation:
\begin{equation}
    N_{send} = \lfloor N_{total}*(P_{min}+k*(100-P_{min})/K)\% \rfloor, k\in\{0, 1, ..., K\}
\end{equation}

The compute node only needs to send the first $N_{send}$ channels of features at the transmission step. Additionally, casting the data type from Float32 to Float16 before transmission also helps to reduce data size.

\section{Experiment}
\subsection{Dataset}
CIFAR-10 and CIFAR-100 are datasets that contain 10 classes and 100 classes of images respectively. Both of them have 50000 training images and 10000 testing images. Every image has a size of 32$\times$32 and RGB three channels. It is more general than MNIST, so we do all the experiments on these datasets. Furthermore, we preprocess the images with Autoaugment \cite{cubuk2018autoaugment}, a bunch of data augmentation policies searching by reinforcement learning, in the training stage. The operators in Autoaugment include shearing, rotation, contrast, brightness, sharpness, etc., and the policies are the combination of operators.10\% of the training images are chosen randomly to be validation data.

\subsection{Implementation details}
First we use stochastic gradient descent (SGD) optimizer to train all the original models for 200 epochs from scratch. The learning rate is initially set as 0.1, with a decay schedule of dividing the learning rate by 10 every 50 epochs. The batch size is fixed to 128. As for SNN, we construct an RNN controller with 100 LSTM cells. In the first stage, we train the separable networks through the entire training dataset. Then we train the controller network through the validation dataset for 500 steps in the second stage. In the third stage, we fix all the parameters and sample 100 architectures. Every architecture makes inference on a mini-batch of the testing dataset and the one with the highest accuracy will be kept. These three stages are repeated for 60 times iteratively. Finally, we fine-tune the best model with a learning rate of 0.0001 for 45 epochs. 

\begin{table*}[h]
\centering
\begin{tabular}{lcc}
\hline
\multicolumn{1}{c}{\multirow{2}*{Method}}            & \multicolumn{2}{c}{Test accuracy} \\ \cline{2-3} 
\multicolumn{1}{c}{}                                   & CIFAR-10          & CIFAR-100        \\ \hline
ResNeXt-56 (4$\times$16d)                                     & \textbf{95.76\%}  & \textbf{79.30\%} \\
Sep-ResNeXt-56 (4$\times$16d, $\alpha=2, G=4$)  & 95.53\%           & 79.10\%          \\ \hline
ResNeXt-110 (8$\times$16d)                                    & 96.23\%           & 81.11\%          \\
Sep-ResNeXt-110 (8$\times$16d, $\alpha=2, G=4$) & \textbf{96.32\%}  & \textbf{81.98\%} \\ \hline
ResNeXt-56 (64$\times$4d)                                     & \textbf{96.30\%}  & 81.35\%          \\
Sep-ResNeXt-56 (64$\times$4d, $\alpha=2, G=4$)  & 96.05\%           & \textbf{81.68\%} \\ \hline
\end{tabular}
\caption{Performance of selected models on CIFAR-10 and CIFAR-100.}
\label{table:cifar10&cifar100}
\end{table*}

\begin{table}[h]
\centering
\begin{tabular}{|c|c|c|c|}
\hline
Choice ID                        & Decisions   & Choice ID & Decisions                            \\ \hline
0                         & 0, 1, 2, 3                        & 12                        & 2, 0, 1, 3                        \\ \hline
1                         & 0, 1, 3, 2                        & {13}  & {2, 0, 3, 1} \\ \hline
2                         & 0, 2, 1, 3                        & 14                        & 2, 1, 0, 3                        \\ \hline
3                         & 0, 2, 3, 1                        & 15                        & 2, 1, 3, 0                        \\ \hline
4                         & 0, 3, 1, 2                        & {\color{red}{16}} & {2, 3, 0, 1} \\ \hline
5                         & 0, 3, 2, 1                        & {\color{red}{17}} & {2, 3, 1, 0} \\ \hline
6                         & 1, 0, 2, 3                        & {\color{red}{18}} & {3, 0, 1, 2} \\ \hline
{\color{red}{7}}  & {1, 0, 3, 2} & 19                        & 3, 0, 2, 1                        \\ \hline
8                         & 1, 2, 0, 3                        & 20                        & 3, 1, 0, 2                        \\ \hline
{\color{red}{9}}  & {1, 2, 3, 0} & 21                        & 3, 1, 2, 0                        \\ \hline
{\color{red}{10}} & {1, 3, 0, 2} & {\color{red}{22}} & {3, 2, 0, 1} \\ \hline
11                        & 1, 3, 2, 0                        & {\color{red}{23}} & {3, 2, 1, 0} \\ \hline
\end{tabular}
\caption{All the choices of communication decisions and their corresponding id when G=4. The order in a decision means the destination node of transmission from node 0, 1, 2 and 3. For example, decision no.7 represents that node 0 sends to node 1, node1 sends to node 0, node 2 sends to node 3 and node 3 sends to node 2. The choices in red are communication-intensive, which means every node has to send data to other node.}
\label{table:policy_id}
\end{table}

\begin{table}[h]
\centering
\begin{tabular}{|c|c|}
\hline
Choice ID & Sparsity \\ \hline
0         & 50.00\%  \\ \hline
1         & 56.25\%  \\ \hline
2         & 62.50\%  \\ \hline
3         & 68.75\%  \\ \hline
4         & 75.00\%  \\ \hline
5         & 81.25\%  \\ \hline
6         & 87.50\%  \\ \hline
7         & 93.75\%  \\ \hline
8         & 100.00\% \\ \hline
\end{tabular}
\caption{All the choices of sparsity decisions and their corresponding id when K=9.}
\label{table:sparsity_id}
\end{table}

\subsection{Searching for high-performance models} \label{enas}
We construct an RNN controller with 100 LSTM cells to learn the policy for sampling communication decisions. The possible number of choices in every transmission step is $4!=24$, when $G=4$. Table \ref{table:policy_id} shows all decisions and their corresponding id. The searching spaces for separable ResNeXt-56 ($\alpha=2, G=4$) and ResNeXt-110 ($\alpha=2, G=4$) are $24^{9}\approx2.64\times10^{12}$ and $24^{18}\approx6.98\times10^{24}$, respectively. Furthermore, Table \ref{table:sparsity_id} lists all the sparsification decisions when $K=9$ if we additionally consider reducing amount of transmission data. The best decisions learned by the controller under different settings are listed in Table \ref{table:policy_found}, and the results compare with original models are shown in Table \ref{table:cifar10&cifar100}, Table \ref{table:enas_performance}. We notice that a well-trained controller tends to sample communication-intensive decisions, which proves that the connections between separated parts are important. However, some decisions which are not communication-intensive are still sampled. These decisions help reduce the loading of transmission because at least one node do not need to transmit data.

\subsection{Reduction of transmission data}
To further reduce the amount of transmission data, we add decisions as described previously to control the sparsity of data. The sparsity level K is set as 9, so there are 9 levels between 50\% and 100\% and we simply use 0$\sim$8 to stand for the nine levels. We also discuss the impact if we cast full floating-point to half floating-point when transmission. The performance is shown in Table \ref{table:enas_performance}. We found that there is only a little accuracy drop with these techniques, and the total amount of transmission data can be reduced to only 14.43\%, compared with the original model.

\begin{table*}[h]
\centering
\begin{tabular}{lcc}
\hline
\multicolumn{1}{c}{{Method}}                & {Acc.} & {Comm. costs} \\
\hline
ResNeXt-56 (4$\times$16d) w/ ring all-reduce                               & 79.30\%                                                                     & 100.00\%                      \\ \hline
\begin{tabular}[c]{@{}l@{}}Sep-ResNeXt-56 (4$\times$16d, $\alpha=2, G=4$)  \end{tabular}                  & 79.10\%                                                                     & 31.48\%                       \\ \hline
\begin{tabular}[c]{@{}l@{}}Sep-ResNeXt-56 (4$\times$16d, $\alpha=2, G=4$) \\  + sparsification ($K=9$)\end{tabular}        & 78.70\%                                                                     & 27.99\%                       \\ \hline
\begin{tabular}[c]{@{}l@{}}Sep-ResNeXt-56 (4$\times$16d, $\alpha=2, G=4$) \\  + sparsification ($K=9$) + Float16\end{tabular} & 78.56\%                                                                     & 14.43\%                       \\ \hline
\end{tabular}
\caption{Performance and transmission benefit under different combinations of techniques on the CIFAR-100 dataset.}
\label{table:enas_performance}
\end{table*}

\begin{table*}[h]
\centering
\begin{tabular}{lc}
\hline
\multicolumn{1}{c}{Method} & Decision 
\\ \hline
ResNeXt-56 (4$\times$16d) & -  
\\ \hline
Sep-ResNeXt-56 & [12, 12, \textcolor{red}{7, 9, 10, 23, 9, 16, 18}]   
\\ \hline
\begin{tabular}[c]{@{}l@{}}Sep-ResNeXt-56 \\ + sparsification\end{tabular}           
& \begin{tabular}[c]{@{}c@{}}{[}{[}0, 8{]}, [\textcolor{red}{10}, 3], [\textcolor{red}{18}, 7], [\textcolor{red}{22}, 8], \\ {[}\textcolor{red}{16}, 8], [\textcolor{red}{18}, 8], [\textcolor{red}{9}, 7], [\textcolor{red}{13}, 8], [\textcolor{red}{22}, 7]]\end{tabular} 
\\ \hline
\begin{tabular}[c]{@{}l@{}}Sep-ResNeXt-56  \\+ sparsification + float16\end{tabular} 
& \begin{tabular}[c]{@{}c@{}}{[}{[}\textcolor{red}{9}, 4{]}, {[}\textcolor{red}{22}, 6{]}, {[}\textcolor{red}{18}, 6{]}, {[}3, 6{]}, \\ {[}\textcolor{red}{16}, 7{]}, {[}\textcolor{red}{23}, 8{]}, {[}\textcolor{red}{9}, 8{]}, {[}\textcolor{red}{13}, 8{]}, {[}20, 7{]}{]}\end{tabular}  
\\ \hline
\end{tabular}
\caption{Best communication decisions sampled by controller under different combinations of techniques. The decisions in red are communication-intensive.}
\label{table:policy_found}
\end{table*}

\subsection{The benefit of the controller in neural architecture search}
\begin{figure*}[h]
\begin{center}
\includegraphics[width=.99\linewidth]{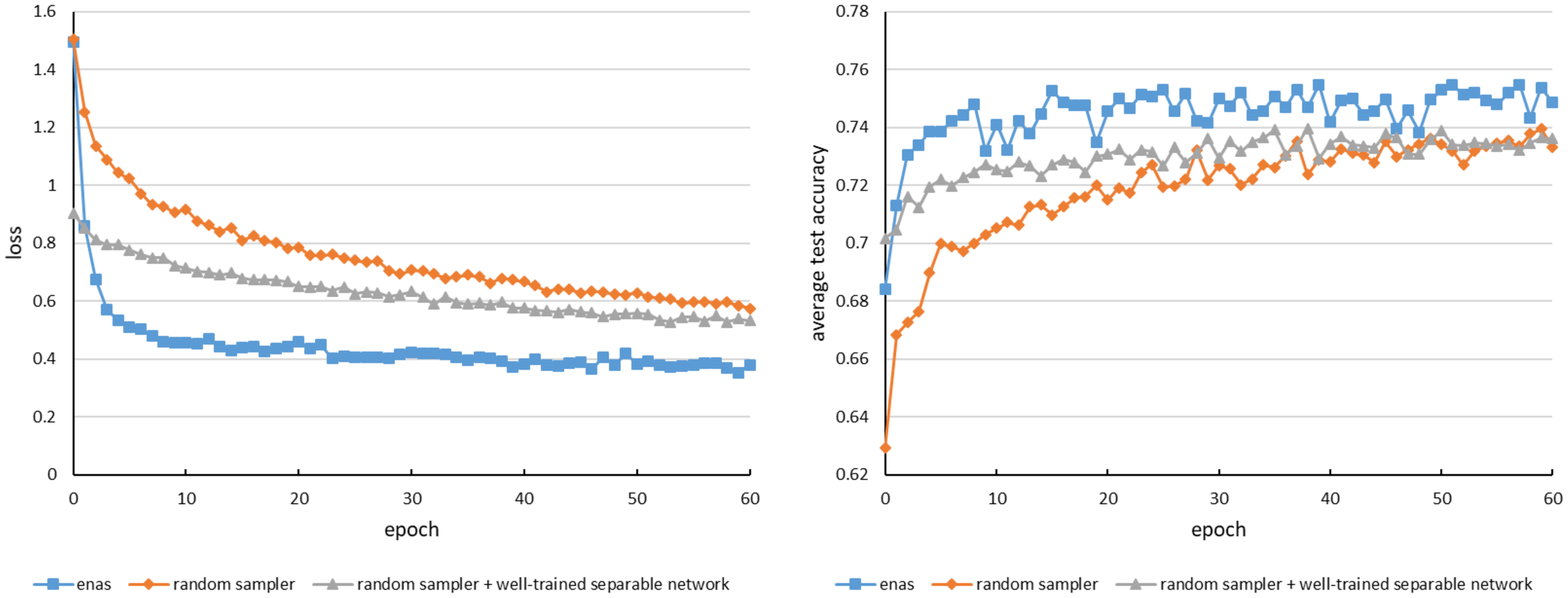}
\end{center}
   \caption{Training curves in different cases. (Blue) Untrained separable network + RNN controller. (Orange) Untrained separable network + random sampler. (Gray) Well-trained separable network + random sampler.}
\label{fig:enas_training_compare}
\end{figure*}

In this section we discuss if the controller really learned to sample better policies. We reproduce the experiment in the previous section by replacing the RNN controller with a random sampler. To reduce the effect of how well a separable network is trained, we also design a control group that uses the pre-train weight from the case with the controller to initialize the separable network. The training loss and average testing accuracy are shown in Fig. \ref{fig:enas_training_compare}. It is clear to see that the case with the controller outperforms cases without a controller, showing that the controller really helps to achieve better decisions.

\begin{figure}[h]
\begin{center}
\includegraphics[width=.99\linewidth]{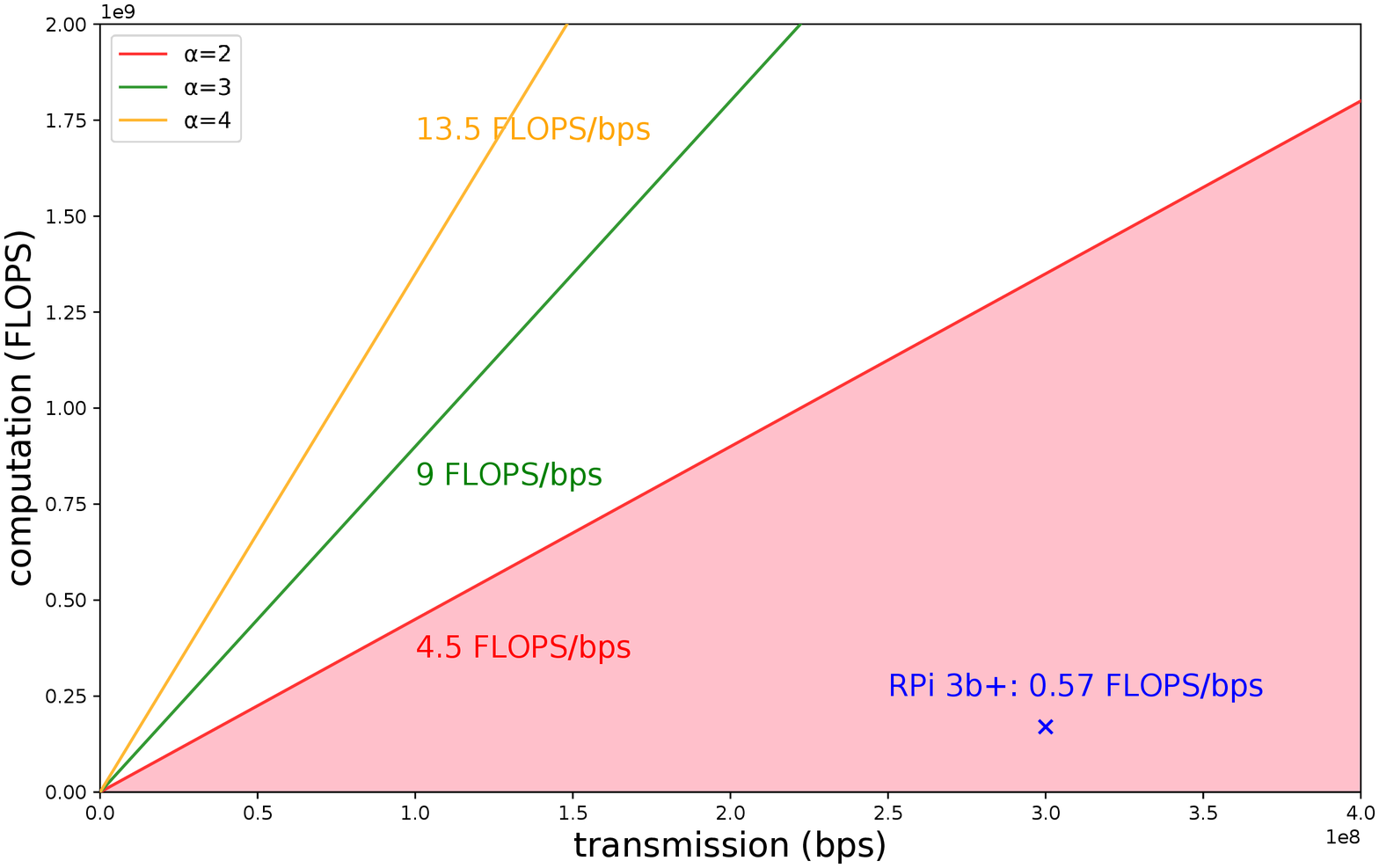}
\end{center}
   \caption{Ratio of computation to transmission under different $\alpha$.}
\label{fig:simulation}
\end{figure}

\subsection{Deployment analysis}
\begin{figure}[h]
\begin{center}
\includegraphics[width=.99\linewidth]{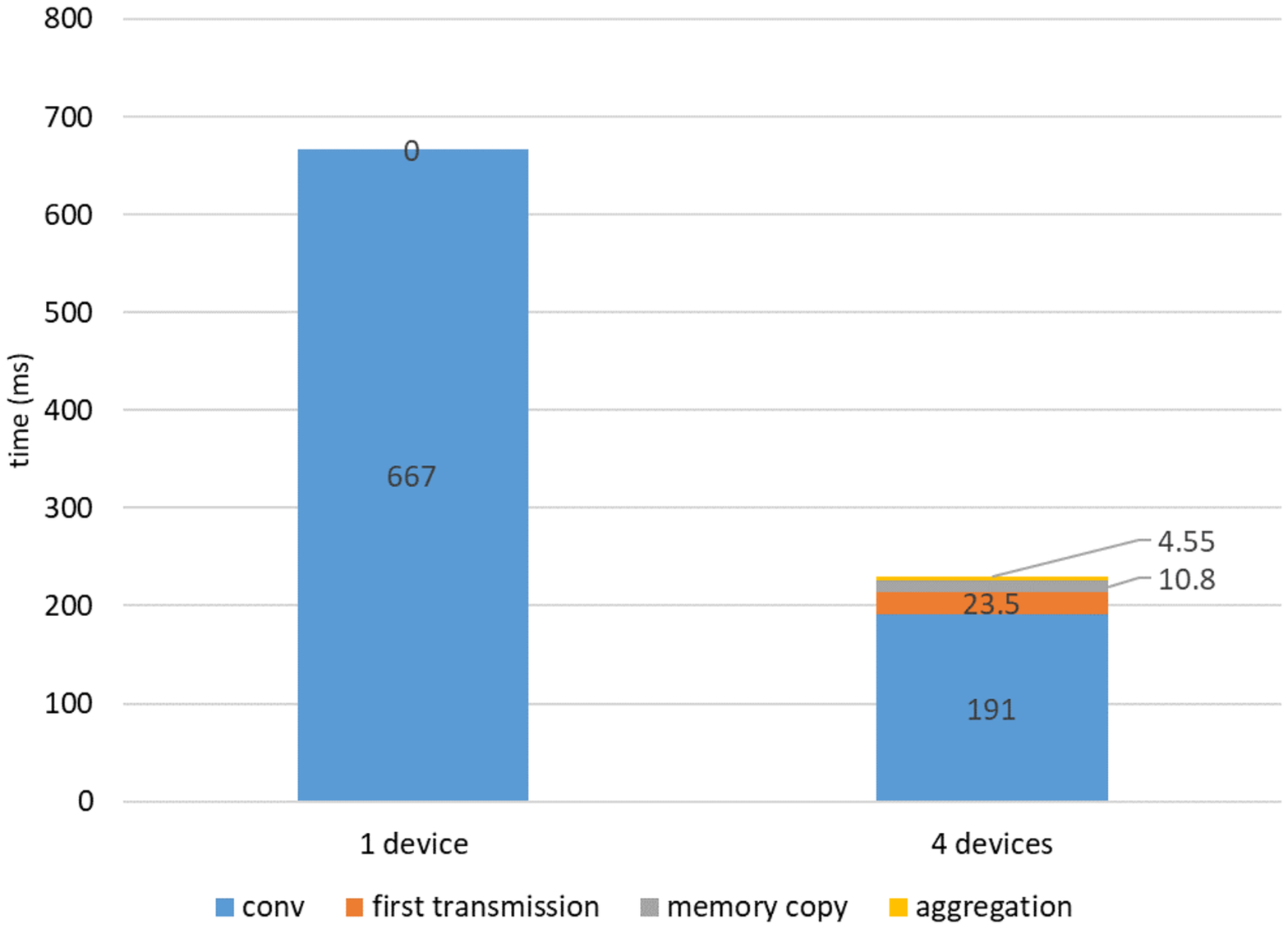}
\end{center}
   \caption{Time consumed by different operations. (Left) Separable neural network deployed on one device. (Right) Separable neural network deployed on four devices. }
\label{fig:tx2_result}
\end{figure}

\hspace{\parindent} To deploy our model on edge clusters, we further specify the relation between our settings and specifications of devices. With appropriate setting of staleness $\alpha$, we can theoretically mitigate transmission overhead. That is, if there is a given separable model, the computation and amount of transmission data are known, and the upper bound of the ratio can be determined. Fig. \ref{fig:simulation} shows the system requirements to deploy separable ResNeXt-56(4$\times$16d, $G=4$) under different $\alpha$. We can see that under same computing power, $\alpha$ would be set larger if the transmission is much slower. The area under the line means that the specifications of the devices can match the setting of $\alpha$. For example, we evaluate our target device Raspberry pi 3b+, which has computing power of $1.7\times10^{8}$ FLOPS and 300Mbps transmission speed. The result indicates that the specification is sufficient to deploy our model. As for the implementation in real scenario, we use 4 devices with quad-core ARM A57 to deploy separable ResNeXt-56(4$\times$16d, $G=4$) model. The time measurement of different components is shown in Fig. \ref{fig:tx2_result}. Although there are some overheads on memory copies, first transmission and aggregation of feature maps, we can speed up inference about 3X. 

\section{Conclusion}
We proposed a new approach of parallelizing machine learning models to enable the deployment on edge clusters and make inference efficiently. Different from traditional parallelism methods, we focus on the reduction of transmission costs through the architecture design to solve the problem of transmission overheads. With our approach, the latency of inference and the model size on one device can be decreased greatly. The communication overheads can be balanced by setting the proposed staleness factor. We also apply the techniques of the neural architecture search to find the best performing models and further reduce transmission. Overall, our work provides a solution to aggregate computing power of edge devices in a cluster. Large models can be deployed properly without significant performance drops. For future research, our work can  combine with model compression techniques to further decrease latency. The support of heterogeneous networks should also be considered to fit the scenario in real life more precisely.

\section*{Acknowledgement}
The work is supported in part by Ministry of Science and Technology, Taiwan, with grant no. 109-2221-E-002-145-MY2.

\bibliography{mybibfile}

\end{document}